\documentclass{elsart}
\usepackage{graphicx}
\usepackage{rotating}
\usepackage{amssymb}

\newcommand{\be}{\begin{equation}}
\newcommand{\ee}{\end{equation}}

\begin{document}
\begin{frontmatter}
\title{ Punctuation effects in english  and  esperanto texts}
\author{M. AUSLOOS  }
\ead{marcel.ausloos@ulg.ac.be}
\address{previously at : GRAPES@SUPRATECS,\\ Universit\'e
de Li\`ege,   Sart-Tilman, \\B-4000 Li\`ege, Euroland
\\{\it nowadays at :
7 rue des Chartreux, B-4122 Plainevaux, Belgium}
}


\begin{abstract}
A statistical physics study of punctuation effects on sentence lengths is presented for written texts: {\it Alice in wonderland} and {\it Through a looking glass}. The translation of the first text into esperanto is also considered as a test for the role of punctuation in defining a style, and for contrasting natural and artificial, but written, languages.
Several log-log plots of the sentence length-rank relationship are presented for the major punctuation marks. Different power laws  are observed with characteristic exponents.
The  exponent can take a value much less than unity ($ca.$ 0.50 or 0.30) depending on how a sentence is defined.   The texts are also mapped into  time series based on the  word frequencies.  The quantitative differences between the original and translated texts are very minutes, at the exponent level. It is argued that sentences seem to be more reliable than word distributions in discussing an author style.

\end{abstract}

 
\end{frontmatter}%

\section{Introduction}

Since  \cite{ebelingsentence}, 
there is a relatively interesting set of studies pertaining to the structure of written texts through techniques based on   statistical physics ideas and methods, usually measuring the word length or/and word frequency distribution. Without claiming to be exhaustive, let us mention recent  studies, much after 2000, on
german  \cite{sentencegerman}, 
polish \cite{polishdrozd}  
 english and irish \cite{polishdrozd,montemurro2606,kassel,sentenceirish,englchinsentence,haenglchinsentence},
chinese \cite{englchinsentence,haenglchinsentence,rousseauchinese,chineseoldnew},
japanese \cite{sentencejapanese},  
greek  \cite{greek1,greek2,LTSpanos},
turkish \cite{turkish},
hungarian \cite{hungarian},
welsh  \cite{welsh}, 
baltic and slavic  \cite{balticslavic},
but also in less natural languages like
 fortran \cite{fortran},
 artificial  \cite{meadowartificial},
or esperanto \cite{Ausloos}.
Of course these studies are partially  a revival of an enormous flurry of studies in linguistics which started as early as  1930  and included later on work by Zipf and many others\cite{Zipf2,yule,powers}.

Debates exist whether  a few texts are  sufficiently representative of a language and how big a  lexicon must be before it  becomes significant. This $caveat$ presented, it is fair to say that it seems that several specific features of written texts have not been studied in detail. The role of punctuation on the structure of texts is one of these.

According to wikipedia the first inscription with punctuation mark is the Mesha Stele (9thBC); see  $http://en.wikipedia.org/wiki/Mesha_{-}Stele$.  A long time ago
Greeks and Romans adopted a few punctuation marks (the dot and combinations, essentially) in order to mark pauses in texts, to be played. Other historical details on the creation, dissemination, use and types of punctuations in various languages can be found in \\  $http://en.wikipedia.org/wiki/Punctuation$, and  \\$http://grammar.ccc.commnet.edu/grammar/marks/marks.htm$. 

Through these e-references, it can be learned that  punctuation marks are symbols that indicate the structure and organization of a written text in a specific language, for readability, as much as for suggesting intonation and pauses when reading aloud. In written English, punctuation is vital to disambiguate the meaning of sentences, though this does not go  without problems  \cite{Ehrlich,rules}.

Notice that some modern writers have attempted to go in some sense backward.      As far as 1895,  Crane published  {\it The Black Riders and Other Lines} \cite{crane} in capital letters: the poems   appearing  without  punctuation, an unusual typographical presentation for the time, -  a style system considered as garbage by the critics. In another language, e.g. french,   Apollinaire  \cite{apollinaire}  published one of his major pieces $Alcools$  without punctuation. Thereafter, Similarly,   the french surrealists and dadaists scorned punctuation, like Aragon  \cite{aragon} who avoided any in most of his poems and prose for/about Elsa Triolet.  That  followed from the para-psychological theory put forward by Breton  \cite{Breton} in {\it The Manifesto}, containing new/practical recipes  for enhancing the {\it Magic Surrealist Art}, such as: "...Punctuation of course necessarily hinders the stream of absolute continuity which preoccupies us ... ".  This was recently "poetically"  reformulated by Hahn \cite{hahn} in {\it The Pity of Punctuation} poem.  Some "maximum" was likely reached by Joyce \cite{joyce}. In $Ulysses$ symbolically  conserving the structure of HomerÕs  {\it The Odyssey}, where there is no punctuation, 
Joyce omits punctuation entirely, in the last chapter of the novel, - consisting of eight long paragraphs, in order to mimic the uninterrupted flow of naked thoughts.

Thus punctuation could be avoided. Indeed there is some redundance, since a capital letter can indicate to the reader a new sentence. 
 One major difficulty nevertheless occurs in text analysis: it is more easy to observe a punctuation sign on a text than a capital letter.

However, fundamentally, in literature, the marks are strongly depending on the writer choice \cite{aragon,Breton,joyce}, but also on the editor \cite{NASA,ulyssesreadersdigest}.  A  question can thus be raised about the relationship, if any, between an author and the use of punctuation marks, for defining his/her style.

A few text studies seem to exist along these lines in the recent literature, i.e.  having considered   structures at the sentence level in english\cite{ebelingsentence,englchinsentence,haenglchinsentence}, in german \cite{sentencegerman}, in chinese \cite{haenglchinsentence}, in  japanese \cite{sentencejapanese},    sometimes strangely neglecting the punctuation role as in  \cite{polishdrozd,sentenceirish}.
In order to propose further studies on the matter, it is attempted here to discuss well known written texts. Moreover, as in \cite{polishdrozd,Ausloos} such considerations are extended to some translation of texts. 

 The texts here below chosen are freely available from the web \cite{Gutenberg}, i.e. {\it Alice in wonderland} (AWL) \cite{carollAWL} and {\it Through a looking glass} (TLG)  \cite{carollTLG}.
They 		are representative  of a  well known mathematician Lewis Caroll.  Such a choice will allow one to discuss whether the differences between two single author english texts, having appeared at different times (1865 and 1871),   contain different structures. The first text is also available in esperanto \cite{esperanto,footn2}.  It seems in order to  observe whether some style or structural change has been made between a text and its translation,   thus  whether the translation observes  similar statistical rules as the original text from the punctuation point of view.  Previous work on  the english AWL version should be here mentioned \cite{powers}.

 In Sect. 2,   the methodology is   briefly exposed. It is recalled   that one can map texts into  (word) length time series (LTS) or into a (word) frequency
time series (FTS). In the present case one adopts the length time series  approach in order to count the number of characters (and blanks) in a sentence, i.e. defining a time interval ending by some punctuation mark	 Some test with a  FTS will be made. In Sect. 3,  the results are presented through log-log plots of the sentence length-rank relationship and  along a Zipf analysis for the word distribution. 
 A conclusion with statistical and linguistics  comments is found  in Sect. 4.

 \section{Data and Methdology}
 
 For the present considerations  two texts here above mentioned   and one translation have been selected and downloaded from a freely available site \cite{Gutenberg}, resulting obviously into three files. The chapter heads  are not considered. All   analyses are carried out over this reduced file  for each text.   As indicated in the introduction, one can  look at the length of sentences, or bits of sentences, taking into account relevant separators: ``.'' (dot) and ``,'' (comma), colon, semi-colon, exclamation point, question mark, i.e.,  ``:'', ``;'', ``!'', ``?''.  
 
  
By analogy with the  original Zipf analysis method or technique which gives a rank $R$ to the words according to their frequency $f$ and make a log-log frequency-rank plot, one ranks here the sentences according to their length $l$ (to be defined) and one searches for $l(R)$.  Usually, for many languages, written texts, one has $ f \sim R^{-\zeta}$, such that one roughly sees a straight line going through the data, o a log-log plot,  interestingly with a slope $\simeq -1$ \cite{west}.  A large set of references on Zipf's law(s) in natural languages can be found in \cite{httpZipf}.  
  Thus, one  considers that there is a one-to-one relationship between rank and frequency.  This is strictly true if there is no ambiguity in the ranking; sometimes  two (or more) words have the same frequency.  Their rank  has been attributed according to their chronological appearance in the text, apparently without much loss in information content.
 
As previously mentioned,  this Zipf law and others have been mainly considered at the word distribution level; it is fair to reemphasize related work  at the sentence level, defined through separation dots,  e.g.  on german \cite{ebelingsentence,sentencegerman}, on irish \cite{sentenceirish,englchinsentence}  and  on  japanese \cite{sentencejapanese}  texts.

\begin{figure}
 \begin{center}
\hspace{-0.2cm}
\includegraphics[width=2.7in] {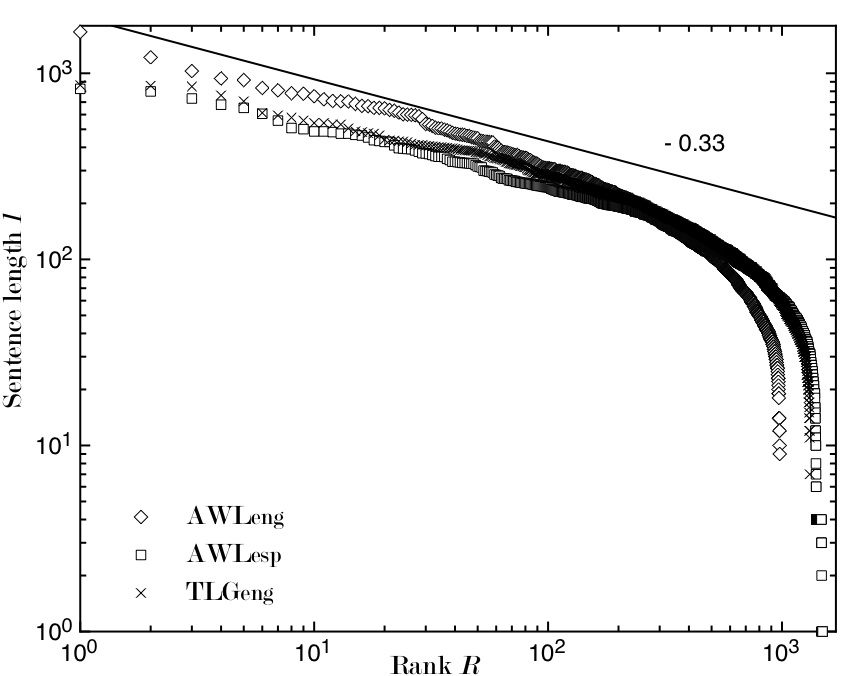}
\vspace{-0.1cm}
\includegraphics[width=2.7in] {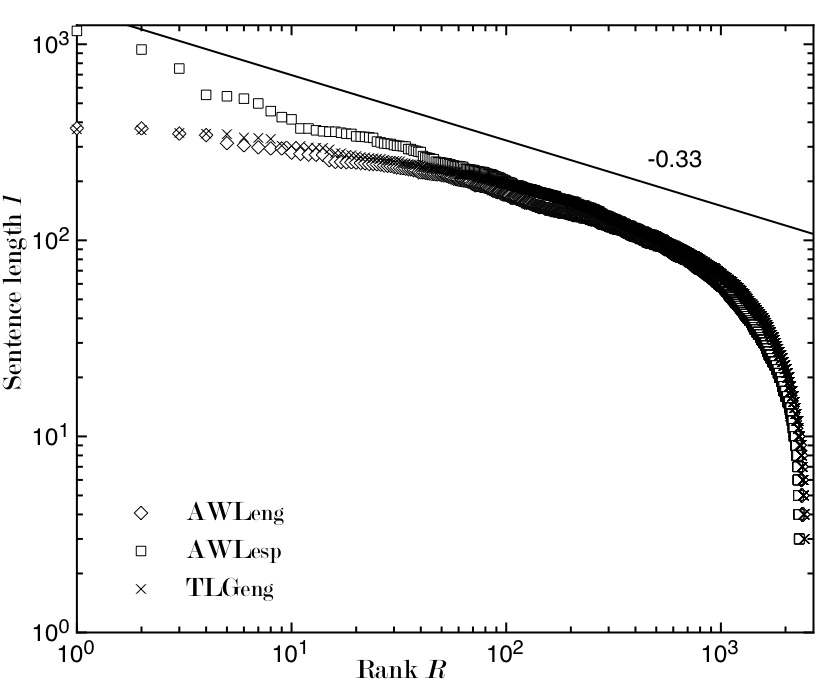}
\vspace{-0.01cm}
\caption{\label{3}  log-log plot of the rank-sentence lengths, as separated by (a) dots, (b) 
commas,  in the three texts of interest AWL$_{\mbox{eng}}$, AWL$_{\mbox{esp}}$ and TLG$_{\mbox{eng}}$. The  $\eta=$ 0.33 exponent of the corresponding rank law is indicated as a guide to the eye}  
\end{center}
\end{figure}

\begin{figure}
 \begin{center}
\hspace{-0.2cm}
\includegraphics[width=2.7in] {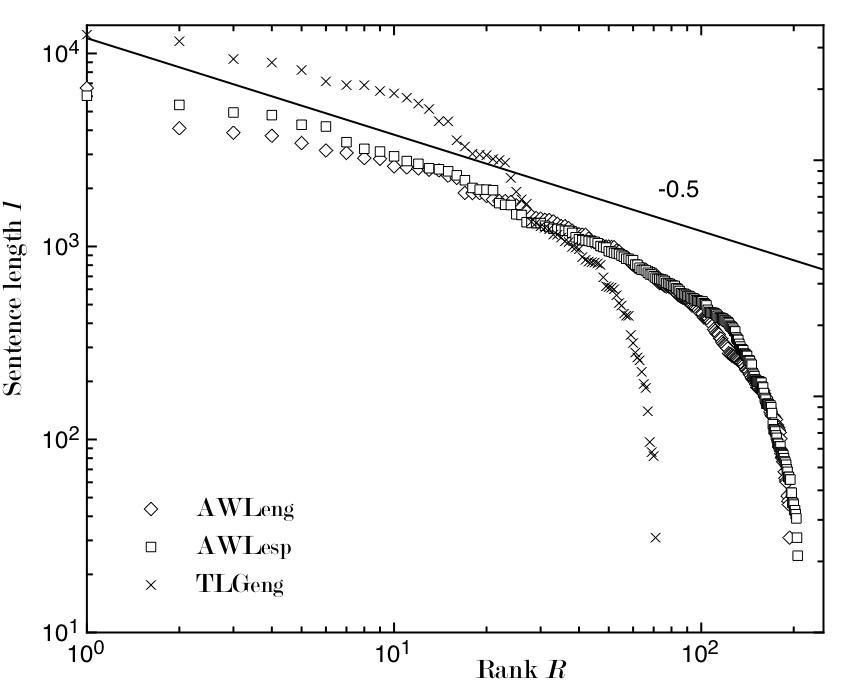}
\vspace{-0.1cm}
\includegraphics[width=2.7in] {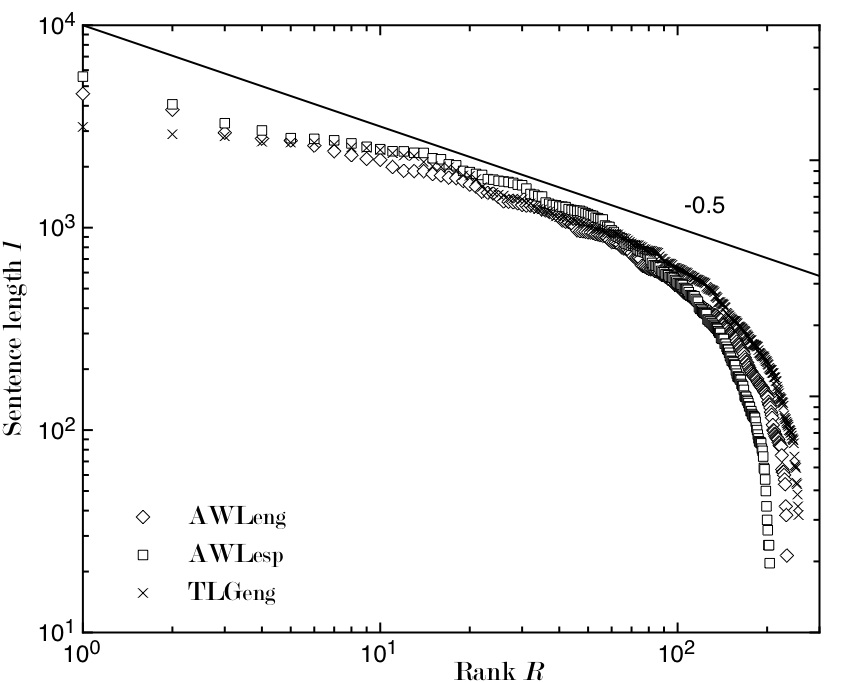}
\vspace{-0.01cm}
\caption{\label{4} log-log plot of the rank-sentence  lengths, as separated by (a) semicolons, (b) colons,  in the three texts of interest AWL$_{\mbox{eng}}$, AWL$_{\mbox{esp}}$ and TLG$_{\mbox{eng}}$. The  $\eta=0.50$ exponent of the corresponding rank law is indicated as a guide to the eye} 
 
\end{center}
\end{figure}

 \begin{figure}
 \begin{center}
\hspace{-0.2cm}
\includegraphics[width=2.5in] {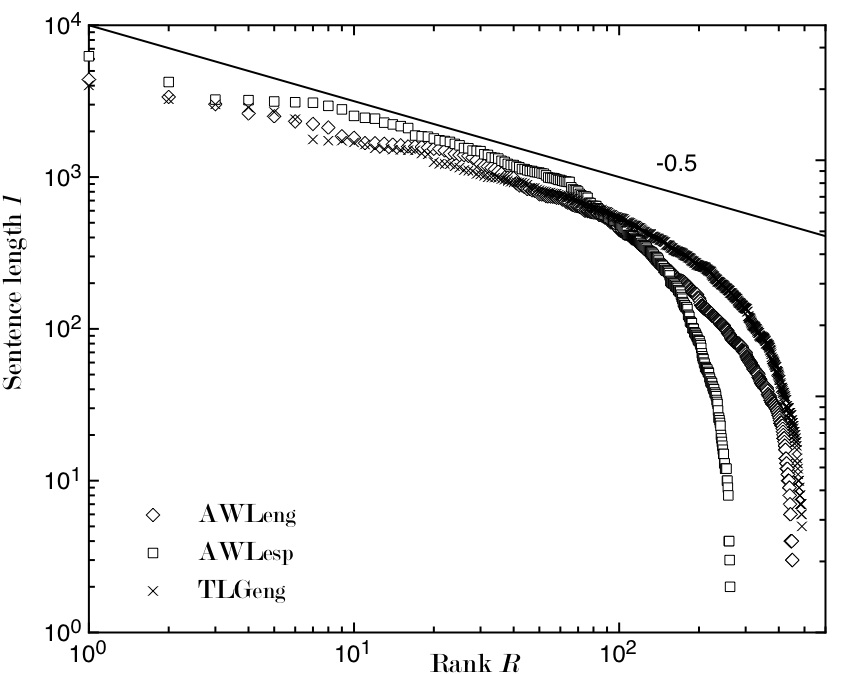}
\vspace{-0.01cm}
\includegraphics[width=2.5in] {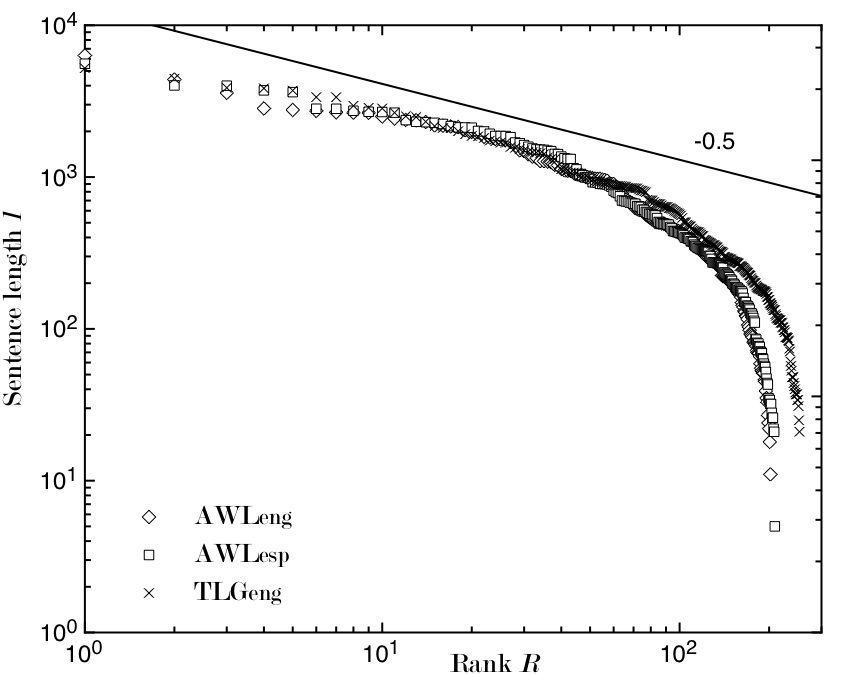}
\caption{\label{5} log-log plot of the rank-sentence  lengths, as separated by  (a) exclamation points, (b) question marks,  
in the three texts of interest AWL$_{\mbox{eng}}$, AWL$_{\mbox{esp}}$ and TLG$_{\mbox{eng}}$. The $\eta=0.50$ exponent of the corresponding rank law is indicated as a guide to the eye } 
 
\end{center}
\end{figure}

 \begin{figure}
 \begin{center}
\hspace{-0.2cm}
\includegraphics[width=2.8in]{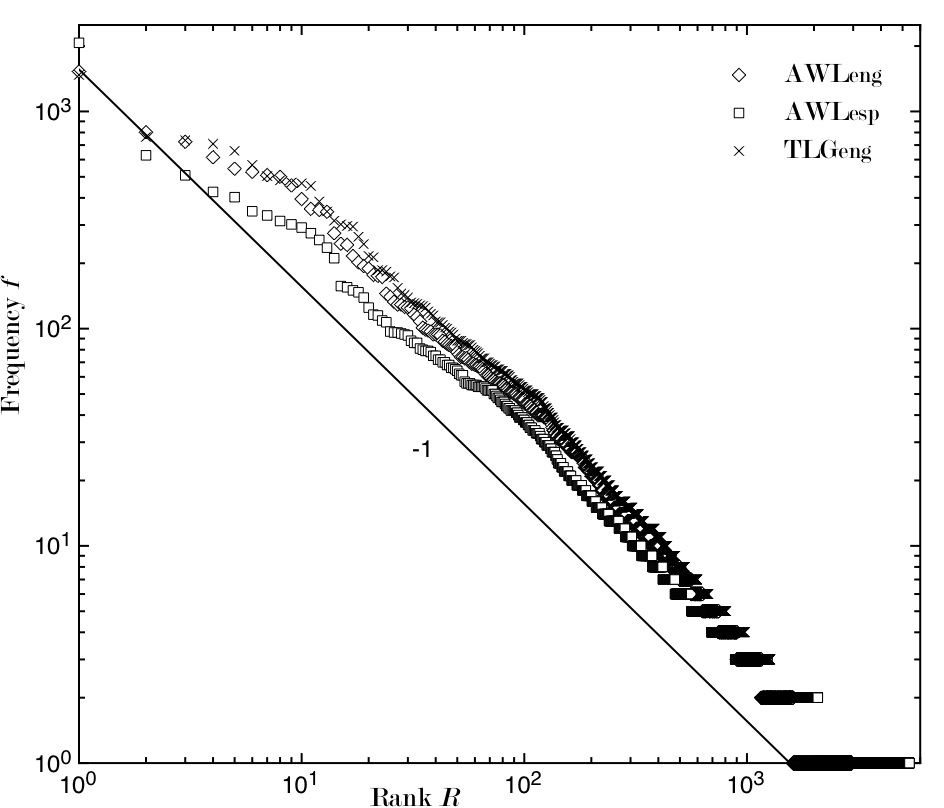}
\vspace{-0.2cm}
\caption{\label{1} Zipf (log-log) plot of the frequency of words in the three texts of interest AWL$_{\mbox{eng}}$, AWL$_{\mbox{esp}}$ and TLG$_{\mbox{eng}}$. The usual ($\zeta=1$) exponent is indicated }
\end{center}
\end{figure}

  \section{Results}

   In the present case, one considers the length $l$ of sentences defined between various punctuation marks, counting characters rather than words.  The punctuation marks  define {\it time intervals} or  bits of $sentences$. 
The result of the LTS rank-like analysis, a length-rank relationship,  for the three main texts is shown in Figs. \ref{3}-\ref{5}, for  the different punctuation marks,  as mentioned in the figure captions.   For pointing out the cases, it is seen that the longest ''sentence''	contains 1669, 825, and 864	 characters, in AWL$_{\mbox{eng}}$, AWL$_{\mbox{esp}}$	and TLG$_{\mbox{eng}}$ respectively, when the sentence ends with a dot, Fig. 1(a). Similarly  the longest sentence	contains	6323	, 5581	and 5212 characters when ending with question mark, in the respective texts.
Several orders  of magnitude in the maximum rank and in the lengths  immediately distinguish the cases.  There about 2000, 200 and 500 ranks, i.e. different lengths,  depending on the punctuation marks, grouped as in Figs. \ref{3}-\ref{5}. On the other hand the  length  can vary much:  from about  300 to 12000  depending on the   cases.

At once it is observed that  TLG$_{\mbox{eng}}$ slightly differs from others,  when  the sentences end with a semi-colon, Fig. \ref{4}(a).     Interestingly let it be observed that the Rank =1 length for the esperanto text is often higher than for the english texts.    This might be argued to originate from the number of available words to make any sentence. The number of words is less in esperanto indicating a greater simplicity in vocabulary: english  contains  ca.   1 M  words \cite{monitor}, esperanto 150 k words \cite{plena}.  
 
 Each log-log plot roughly  indicates a simple power law relationship, for $ca.$  $R\le500$, $R\leq50$, $R\le100$, i.e. $l    \sim R^{-\eta}$.
This corresponds to a    break length  value  $ca.$ 100, 1000 and 1000. Some curvature is found for all texts below $R\sim5$ where a so called discontinuity exists. It can be understood in lines of comments by \cite{powers} on word frequency plots. For the latter case,  this is due to  a transition between colloquial (''common'') small  and ''distinctive'' words; one can be easily convinced of the analogy when  forming and studying sentences.  This weak change in curvature at low rank value  feature is also  explained  in
 Mandelbrot \cite{3,bbm2,bbm1} using arguments based on
   $fractal$ ideas, applied to the structure of lexical trees. Some marked break, or change in slope,  looking like a distribution truncation is also found for $R$ large. Some discussion  for the latter case in discussing word distributions can be read in \cite{powers}.  By analogy this behavior is thought to arise from the scarcity of long sentences, i.e. there is much difference in the number of characters for the long sentences, not so much for the small ones. In some physics-like sense one would attribute the result to the  polycrystalline nature of the sample, made of few big crystals and many tiny ones.


The most drastic difference occurs between the cases
of the first group of punctuation marks, Fig. \ref{3},  where the slope indicates that the exponent $\eta$ is  rather close to 1/3,  and  the latest four cases, Fig. \ref{4}-\ref{5}, where  the slope is closer to 1/2.  Notice that the number of punctuation marks is relatively equivalent 
in all texts, as estimated from the integral of the distributions \cite{footnote1}, but there are many more  dots and commas   than   other punctuation marks,  as is expected indeed, - a factor of ten. Therefore one might expect some stronger finite size effect  influencing the exponent value in the latter cases.

In conclusion of this section, let it be accepted that the quality of the power-law fits   are   less impressive in the case of the length of sentences (Figs. 1-3) than in the case of the frequency of words (Fig.4). This observation possibly indicates that the existence of a cut-off is more likely, for the length of sentences, because of grammatical and readability constraints. One could alternatively present the data in a log-normal plot,  and observe whether a stretched exponential can be considered. However the shortness of the data range is in this case a handicap as well.
 Further "theoretical" discussion on the values of the exponents as found below  in Sec. \ref{sec:concl} would not be more agreeable. The possible stretched exponential is thus below briefly discussed.

\section{Conclusion} \label{sec:concl}
 
 The occurrence of  such a power law  for word distributions has already been suggested \cite{zipfregimes} to originate in the ''hierarchical structure'' of  texts as well as to the presence of long range correlations (sentences, and logical structures therein).
  Some {\it ad hoc}  thought has been presented based on constrained correlations \cite{modelzipf1,modelzipf2}.  
A value of $\eta$ smaller than unity indicates a wide, flat distribution thus a more homogeneous repartition of the variables (lengths of sentences, here).  
 
Gabaix \cite{gabaix}, looking at city growths,  claims that  two causes can lead to a value less than 1.0, i.e.  either  (i)  the growth process deviates from Gibrat's law \cite{gibrat} which assumes that  the mean  growth rate  is independent of the size,
 or/and
 (ii)   the variance of the growth process   is size-dependent.
 Recall that one does not examine the ''growth'' of the text at this stage yet, nor have any model for  doing so presently.

However, one can imagine the way L. Caroll  (and other authors) function. After writing a first draft of some chapter, the author adds, removes, modifies words and sentences, introducing different "grams" leading  to   a modified story development and text structure. Modifying again the text after a second reading, etc. The process is kinetic indeed and basically a growth process, somewhat similar to city growth; Thus it is {\it a priori} hard to say whether  the causes (i) or (ii) or both are influencing the exponent values.

 One can nevertheless debate whether   the sample size is relevant for  estimating a (small) $\eta$ value on so few rank decades \cite{footnote1}.  The same can be considered if the data would seem to fit a stretched exponential. If so it might be argued that an external constraint must be envisaged, as if the writer was influenced by e.g. the size of the paper sheet on which he/she is writing. The present author wishes not to enter into such considerations, though further studies might be of interest nowadays as when studying blogs and RSS \cite{blogs}  and other (electronic or not) reports which are  strictly limited in size or in the number of allowed characters.  

According to a widespread conception, quantitative linguistics will eventually be able to explain such empirical quantitative findings (such as Zipf  law) by deriving them from highly general stochastic linguistic ÔlawsÕ that are assumed to be part of a general theory of human language \cite{chomsky,Best1999} for a summary of possible theoretical positions). In \cite{meyer}, Meyer argues that on close inspection such claims turn out to be highly problematic, both on linguistic and on science-theoretical grounds.	It has also been argued that it is possible to discriminate between human writings \cite{vilenski} and stochastic versions of texts 
precisely by looking at statistical properties of words.  In contrast  I argue here above that this statement can be extended to sentence statistics.  

 The meaning of the results is  admittedly still somewhat elusive, even though the length distribution of text segments between certain types of punctuation marks is new empirical data. It is fair to mention here a reviewer suggestion \cite{reviewer3}  encouraging  investigations of the length distribution of symbol sequences commonly regarded as a 'unit of thought' or a proper sequence, that is not distinguishing between periods, semicolons, question and exclamation marks.  However to test if such statistical measures can indeed be used to classify a text, e.g., to distinguish authorship, a much larger set of texts should be used. Similarly, from this single example the similarity between the Esperanto translation and the original text may not point to the quality of translation, since perhaps any natural text exhibit similar frequency distribution \cite{reviewer3}, and might be due to other external constraints,  as hinted at the end of a previous paragraph.

Last but not least as on comparing AWL$_{\mbox{eng}}$,  AWL$_{\mbox{esp}}$ , and TLG$_{\mbox{eng}}$,   it seems that the texts are qualitatively similar, which indicates ... the quality of the translator. In this spirit, it would be interesting to compare with results originating from text obtained through  a machine translation, as recently  studied in  \cite{koutsoudas}. It is of huge interest to see whether a machine is {\it more flexible} with vocabulary and grammar than a human translator, - see also \cite{kanter}!

Finally, in summary, it is sufficient here to stress that punctuation marks are an essential part and a long lasting feature of indo-european languages, with a great variety of signs and in their use.  
  At first sight, a time series of a single variable appears to provide a limited amount of information if texts and authorships.   FTS   and LTS result from a dynamical process, which is usually first characterized by its fractal dimension. 
  The first approach should contain a mere statistical analysis of the output, as done through a Rank-like analysis here. It has been found that analytical forms, like  power laws   with different characteristic exponents for the ranking properties exist.  The  exponent can take values $ca.$ 1.0, 0.50 or 0.30, depending on how a sentence is defined.  This non-universality, or even another law,  could be   further examined in order to find whether  there is  a measure of the author $style$ hidden in such statistics/fits.  Moreover one on-going challenge is to sort out the laws of sentence statistics in  texts, written or produced by many authors, like scientific papers, thereby discriminating the percentage of truly personal contribution {\it in the writing}. Another apparently more simple investigation  which is in direct line with previously mentioned  studies \cite{blogs}  is the characterization of sentence statistics in online dynamic media, such as Blogs or RSS feeds, which are usually single author  texts.
 
\bigskip{}

  {\it \bf Acknowledgements}

 This paper results from the work of Mr.  Jeremie  Gillet, when an undergraduate student attending my lectures on fractals.  
 I am very  thankful to JG to allow me to publish the results of his data analysis and for many comments. 
Thanks also to Ms.
Sophie Pirotte for critical comments and warnings.

Thanks are also due to FNRS FC 4458 - 1.5.276.07 project having allowed some stay at CREA and U. Tuscia and to the COST Action MP0801 in particular  through the STSM 5378 for financial support at BAS, Sofia,  thus at various stages of this work.


\begin{thebibliography}{99}
\bibitem{footnote1} the number of ''sentences'',  whatever their definition through punctuation marks is: 	1 633, 	2 016 	and	2 059;  the number of characters is: 144 927, 154 445	and 164 147; the number of punctuation marks	is: 4 481, 4 752 	and 4 828.
\end{thebibliography}
\end{document}